\documentclass[11pt]{article}

\usepackage{algorithm}
\usepackage{algorithmic}
\usepackage{agenticlearning-xelatex}
\usepackage{booktabs}
\usepackage{graphicx}
\usepackage{amsmath}
\usepackage{multirow}

\makeatletter
\newcommand{\blfootnote}[1]{%
  \begingroup
  \renewcommand\@makefntext[1]{##1}
  \footnotetext{#1}%
  \endgroup
}
\makeatother

\graphicspath{{figures/}}

\begin{document}

\shorttitle{Vision models and the urban brain}
\shortauthor{Tan \& Deng}

\title{Vision Models Predict Urban Scene Appraisal\\
with Limited Neural Alignment}
\author{
  Kaizhen Tan$^{1}$, Yuantao Deng$^{1}$ \\
  $^{1}$New York University
}
\date{}
\maketitle

\begin{abstract}
Pretrained vision embeddings are widely used to model how people appraise urban
scenes, and they are validated almost entirely by how well they predict human
ratings. Accurate prediction shows that an embedding contains the information
needed to recover the ratings. It does not show that the embedding arranges
scenes as the human visual system does, which is assumed when distances or
dimensions in the embedding are read as perceptual. Using openly released EEG
recorded while 63 adults viewed and rated street scenes of Berlin, we compared
the neural representational geometry of the scenes with the geometry of
pretrained vision models that differ in training objective and size, of simple
image descriptors and of the ratings themselves, each relative to a noise
ceiling given by the agreement between participants. No feature space reached
more than about half of the noise ceiling, which corresponds to about a fifth of
the reliable variance in the neural geometry. A descriptor of oriented
edge energy reached the level of most pretrained models while capturing a
different part of the neural geometry, and deeper layers corresponded to later
neural responses. The same embeddings predicted held-out ratings well, up to
$r=0.87$, yet prediction accuracy and neural correspondence were not reliably
related across models, and the best predictor was among the least aligned.
Predicting how a street is appraised is therefore weak evidence that a model
represents the street as the brain does.
\end{abstract}

\section{Introduction}
\label{sec:intro}

Street-level imagery has become a standard instrument for measuring cities.
Crowdsourced ratings of how streetscapes are perceived can be extrapolated to
whole cities by training a model to predict them from the image
\citep{naik2014streetscore,dubey2016deep}, and the resulting scores are now used
across urban research: reviews of the field count hundreds of studies spanning
transport, greenery, socioeconomic mapping, safety and health
\citep{biljecki2021street,zhang2024urban}. As the method has matured, the image
representation has shifted from engineered descriptors to pretrained deep
embeddings, which are used as general-purpose encodings of what a street looks
like. The ratings themselves are evaluative judgements, such as how safe or how
beautiful a street appears. The urban analytics literature usually refers to them
as urban perception. We call them appraisal, following
\citet{zaehme2026perception}, and reserve perception for the visual processing of
a scene that precedes the judgement.

Validation in this literature is almost entirely predictive: a representation is
accepted when a model built on it recovers held-out human ratings. This tests the
output of the model. It does not test whether the representation arranges
streets as the human visual system does, because the same ratings can be
recovered from representations that arrange the scenes quite differently. The
difference matters once an embedding is used for more than prediction.
Embeddings are also used to group streets into visual types or to compare
neighbourhoods and cities by the distance between their embeddings
\citep{wang2020urban2vec,huang2025measuring,guo2025urban}, and individual
embedding dimensions are interpreted as visual qualities of places
\citep{law2019unsupervised,wang2020urban2vec}. When these distances or dimensions
are read as describing how places look to people, the reading presupposes that
scenes close together in the embedding are scenes that viewers also treat as
alike. Predictive accuracy does not test this presupposition.

One way to test it is with brain recordings, which measure the response to a
scene without requiring an explicit judgement. Representational similarity
analysis compares two systems that do not share a format, such as an EEG
recording and a model embedding, by describing each through the pairwise
dissimilarities among the same set of stimuli and then comparing the two sets of
dissimilarities \citep{kriegeskorte2008representational}. We refer to such a set
of pairwise dissimilarities as the \emph{representational geometry} of the
scenes, and to the agreement between a model geometry and the neural geometry as
their \emph{correspondence}. Because neural recordings are noisy, correspondence
is judged against a \emph{noise ceiling}, the level of agreement between
participants, which is the highest value a model could be expected to reach.
Applied across time to MEG or EEG, the method shows how a scene representation
unfolds, from low-level visual structure early in the response to spatial layout
by roughly 250\,ms \citep{cichy2016comparison,cichy2017dynamics}. Correspondence
between deep networks and visual cortex has since become a benchmark for vision
models \citep{yamins2014performance,schrimpf2018brainscore}, supported by EEG
datasets collected for this purpose \citep{gifford2022large}.

These benchmarks are built around object recognition, whereas urban applications
ask a representation to support evaluative judgements of whole scenes. EEG
studies of urban environments have addressed a different question. They relate
neural responses to the content of a scene, most often by contrasting natural
with built settings \citep{grassini2019processing,mahamane2020natural}. In the
dataset analysed here, the occipital P1, an early component of the visual
response, increases with the ratio of greenery to built area, and the N1 is more
pronounced for scenes with denser straight edges. The ratings are not associated
with these early components, and for some of the scales they are associated with
the later parietal P3 and late positive potential (LPP)
\citep{zaehme2026perception}. In these studies a scene is described by a small
number of hand-specified properties. Whether the learned representations used in
urban analytics resemble the neural representation of a street has not been
measured.

We ask three questions. First, how closely do current vision representations
correspond to the neural representational geometry of urban scenes, relative to
the noise ceiling? Second, how does this correspondence vary with the training objective,
training data, size and layer depth of a model? Third, do models that correspond
more closely to the neural geometry also predict human appraisal better? We
address these questions with the openly released EEG of 63 adults who viewed and
rated 56 Berlin street scenes \citep{zaehme2026perception}, and compare the
neural geometry of the scenes with seventeen feature spaces. Thirteen are
embeddings from pretrained models, selected so that training objective, training
data, architecture and size can each be compared with the others held as constant
as possible. Two are low-level descriptors: Gabor energy, a classical model of
early visual cortex that measures the strength of oriented edges, and colour and
luminance statistics. One is the area proportions of segmented scene elements, a
description widely used in urban studies. These three serve as reference points
for the learned representations. The seventeenth is the appraisal ratings
themselves. In
the experiment every scene was viewed nine times, each time before a different
rating. \citet{zaehme2026perception} averaged over these presentations and left
the effect of the rating goal for future work. Because a vision model returns one
embedding per image whatever the viewer intends, we also test whether the neural
geometry of a scene depends on the judgement the viewer is about to make.

The study makes three contributions.

\begin{itemize}
\item It provides a time-resolved comparison between the vision representations
      used in urban analytics and the neural geometry of urban scenes, scaled by a
      noise ceiling. The comparison is built entirely from public data and
      involves no model training.
\item It shows that the best-aligned vision representation reaches 51\% of the
      noise ceiling, and that a Gabor energy descriptor reaches the level of most
      pretrained models while capturing a largely different part of the neural
      geometry. Larger models are not better aligned in the two model families
      with matched size variants, although deeper layers correspond to later
      neural responses.
\item It shows that accurate prediction of human appraisal does not imply neural
      correspondence. Feature spaces that predict the ratings at $r \ge 0.80$
      reach between 21\% and 51\% of the noise ceiling, and across models the
      two measures are not reliably related.
\end{itemize}

\section{Methods}
\label{sec:methods}

\subsection{Participants, stimuli and task}
\label{sec:data}

We analysed the openly released \emph{Urban Appraisal} dataset (OpenNeuro
\texttt{ds006850}; \citealp{zaehme2026perception}), which contains 64-channel EEG
from 63 adults (mean age 29.2 years, SD 7.5, range 19--61; 33 female, 29 male,
one non-binary) who viewed street-level photographs of Berlin. The original study
was approved by the Ethics Board of the Institute of Psychology and Ergonomics at
Technische Universit\"at Berlin. The present work is a secondary analysis of the
anonymised public release.

The stimuli are 56 screenshots from Apple Look Around street-level imagery of
Berlin. They are a subset of a 100-image set that \citet{sander2024beyond}
assembled by crossing building height (more than five storeys, or five or fewer)
with greenery (present or absent), from which \citet{zaehme2026perception}
retained the most extremely rated images. Each of the four combinations
contributes 14 scenes. The scenes were divided into two blocks of 28, and every
participant saw the same scenes in each block, so that all nine presentations of
a scene fell in the same block. One of the 56 images (\texttt{HTHB08}) is not included in
the public release, so all analyses that require image features use the remaining
55 scenes, and behavioural analyses use all 56. In the segmentation maps released
with the dataset, vegetation covers between 0\% and 52\% of these 55 scenes and
is effectively absent from 26 of them, and buildings cover between 9\% and 79\%.

Each scene was presented nine times to every participant, once for each of nine
appraisal scales, giving 504 trials per participant and 31{,}752 in total. On
every trial a word pair naming the upcoming scale appeared for 1000\,ms, followed
by a fixation period of about 1000\,ms (Appendix~\ref{app:preproc}), the scene for
3000\,ms, and then the rating scale until the participant responded. Three scales
used the nine-point Self-Assessment Manikin (arousal, anchored
\emph{excited--calm}; valence, \emph{happy--unhappy}; dominance) and six used
five-point Likert scales (stress, openness, safety, beauty, hominess,
fascination). Because the arousal and valence anchors run opposite to their usual
direction, we refer to these two scales as calmness and unpleasantness when
reporting rating values.

\subsection{EEG preprocessing}
\label{sec:preproc}

Recordings were sampled at 500\,Hz with FCz as the online reference. The two
blocks are stored as separate recordings and were preprocessed separately. All
preprocessing was done in MNE-Python \citep{gramfort2013meg}. The data were
band-pass filtered between 0.1 and 40\,Hz. A channel was marked as bad when it
was flat, when its largest absolute correlation with any other channel was below
0.4, or when its standard deviation exceeded eight times the montage median, with
at most 10\% of the montage marked per participant (Appendix~\ref{app:preproc}).
Bad channels were interpolated with spherical splines. Ocular components were
removed with ICA (Picard, 30 components), using Fp1 and Fp2 as proxies for the
electro-oculogram because no ocular channel was recorded. FCz was then restored
and the data were re-referenced to the common average.

Epochs were extracted from $-200$ to $1000$\,ms around the stimulus marker,
baseline-corrected on the pre-stimulus interval, resampled to 250\,Hz, and
rejected when any channel exceeded a peak-to-peak range of 200\,\textmu V. This
retained 94.0\% of trials (29{,}838 epochs), with 1.4 channels interpolated per
participant on average. Each epoch carries the identity of the scene, the scale
named in the prompt and the rating given afterwards.

All latencies are reported relative to the stimulus marker in the released event
files. The analysis code of the original authors
(\texttt{perception2appraisal\_analyses} repository) places the ERP windows
60\,ms after the latencies given in their report, to allow for a set-up latency
between the marker and the appearance of the image. We used the same
marker-relative windows for the ERP components (P1 145--185\,ms, N1
192--232\,ms, P3 399--449\,ms, LPP 545--645\,ms). With this reference the
occipital P1 peaks at 164\,ms, compared with about 105\,ms after image onset in
the original report \citep{zaehme2026perception}. We did not shift the data, so
the latencies reported below are about 60\,ms later than latencies relative to
image onset. The offset is the same for all scenes and all feature spaces.

Two participants retained fewer than 60\% of their trials, so that some scenes
had too few trials to enter every cross-validation fold of the dissimilarity
estimate (Section~\ref{sec:geometry}). They were excluded from the
representational analyses, which leaves 61 participants with all 55 scenes and
all 1485 scene pairs. Appendix~\ref{app:robust} reports the model comparison with
these two participants included.

\subsection{Neural scene geometry and noise ceiling}
\label{sec:geometry}

We describe the neural response to the scenes by its representational geometry.
For each participant and time point we computed a $55\times55$ representational
dissimilarity matrix (RDM), in which each entry is the cross-validated squared
Mahalanobis (crossnobis) distance between the scalp patterns evoked by two scenes
\citep{walther2016reliability,guggenmos2018multivariate}. Patterns were first
whitened by multivariate noise normalisation: the within-condition residual
covariance was estimated at each time point with Ledoit--Wolf shrinkage
\citep{ledoit2004well} and averaged over time. The trials of each scene, pooled
over the nine scales, were split into three folds, and the dissimilarity was
estimated as
\begin{equation}
d^{2}_{ij} \;=\;
\frac{\big\lVert \sum_{k}(\mathbf{m}^{k}_{i}-\mathbf{m}^{k}_{j})\big\rVert^{2}
      -\sum_{k}\big\lVert \mathbf{m}^{k}_{i}-\mathbf{m}^{k}_{j}\big\rVert^{2}}
     {K(K-1)},
\label{eq:crossnobis}
\end{equation}
where $\mathbf{m}^{k}_{i}$ is the mean whitened pattern for scene $i$ in fold $k$
and $K=3$. The estimator is unbiased: its expected value is zero when two scenes
evoke the same pattern, so dissimilarities can be compared across participants
and latencies (Appendix~\ref{app:crossnobis}).

Each scene was shown in only one block, and every participant saw the same scenes
in each block. A response component shared by all scenes of a block would
therefore make scenes from different blocks more dissimilar than scenes from the
same block in every participant alike, and it would enter the noise ceiling
without reflecting the images. Within each fold, we therefore subtracted the mean pattern
of each block from the trials of that block before computing the distances,
which leaves the distances between scenes of the same block unchanged. Without
this step, an RDM that only separates the two blocks corresponded to the neural
geometry between 150 and 500\,ms at $\rho=0.22$, 78\% of the lower bound of the
noise ceiling; with it, at $\rho=-0.01$ (Appendix~\ref{app:robust}).

The analyses use two temporal resolutions. Time-resolved analyses use the RDM at
each time point. The main comparison between feature spaces uses a single
window-level RDM, obtained by averaging the neural dissimilarities between 150 and
500\,ms. The window starts where reliable scene structure first appears in these
data (Section~\ref{sec:res-reliable}). It ends at about the time when the neural
geometry settles into the configuration that it keeps for the rest of the epoch,
which is also when exploratory eye movements begin in a substantial share of
trials (Appendix~\ref{app:window}). Appendix~\ref{app:robust} reports a shorter
and a later window. Averaging
reduces noise, so the window-level RDM has a higher noise ceiling than any single
time point within the window.

The noise ceiling was estimated following \citet{nili2014toolbox}, at the same
temporal resolution as the analysis it accompanies. Its upper bound is the mean
correlation between each participant's RDM and the mean RDM of all participants,
and its lower bound is the mean correlation between each participant's RDM and
the mean RDM of the other participants. The upper bound is inflated when the data
are noisy, because each participant contributes to the mean
(Appendix~\ref{app:ceiling}). We therefore express correspondence as a
percentage of the lower bound.

\subsection{Vision representations}

We compared seventeen feature spaces, which we refer to collectively as the
\emph{panel} (Table~\ref{tab:models}). Five are language-supervised: CLIP
ViT-B/32 and ViT-L/14 \citep{radford2021learning}, a CLIP ViT-B/32 trained on
LAION-2B \citep{schuhmann2022laion}, and SigLIP B/16 and SO400M
\citep{zhai2023sigmoid}. Three are self-supervised: DINO ViT-B/16
\citep{caron2021emerging} and DINOv2 ViT-B and ViT-L \citep{oquab2024dinov2}.
Three are category-supervised: ViT-B/16 pretrained on ImageNet-21k and fine-tuned
on ImageNet-1k \citep{dosovitskiy2021image}, ResNet-50 trained on ImageNet
\citep{he2016deep}, and ResNet-50 trained on Places365 \citep{zhou2018places}, the
closest available model trained on scene photographs. Three are dense-prediction
models or explicit scene descriptions: SegFormer-B5 fine-tuned on ADE20K
\citep{xie2021segformer}, the area proportions of the ADE20K categories in the
segmentation maps released with the dataset \citep{zhou2017scene}, and Depth
Anything V2 \citep{yang2024depth}. Two are interpretable low-level descriptors:
Gabor energy over four scales, eight orientations and a $4\times4$ spatial grid,
and colour and luminance statistics. The seventeenth is the group-mean appraisal
ratings. It is not computed from the image. It describes how people evaluate the
scenes, and its correspondence with the neural geometry indicates how far the
neural responses are organised by appraisal. Its RDM is the correlation distance
between the rating profiles of two scenes over the nine scales, computed on the
ratings as coded in the dataset. This distance depends on the range and direction
of each scale, so we also report the Euclidean distance between ratings
standardised across scenes, which does not.

For transformer encoders we extracted class-token activations at four evenly
spaced depths and the projected embedding. For ResNets we extracted spatially
pooled activations at each of the four stages and the logits. For SegFormer we
used the pooled final encoder stage and the predicted class probabilities pooled
to an $8\times8$ grid (its layout), and for Depth Anything the predicted depth map
pooled to a $16\times16$ grid. Model RDMs used correlation distance. Where one
value per model is reported, it refers to the layer whose RDM best matches the
group-mean neural RDM. Selecting the layer on held-out participants gives nearly
the same result (Appendix~\ref{app:selection}).

\subsection{Model--brain correspondence and temporal analysis}

Correspondence is the Spearman correlation between a model RDM and a
participant's neural RDM, averaged over participants. Its uncertainty has two
sources, the sample of participants and the sample of scenes, and we assessed
them separately. Across participants, time courses were tested with a
cluster-based permutation test (2000 sign flips, cluster-forming threshold
$p<0.05$, cluster-mass statistic; \citealp{maris2007nonparametric}), and pairs of
models were compared with Wilcoxon signed-rank tests. Across scenes, we resampled
the 55 scenes with replacement (2000 resamples), recomputed the neural and model
RDMs on each resampled set, discarded the pairs formed by duplicated scenes, and
recalculated the participant-averaged correspondence. Inference over participants
asks whether a result would hold for new viewers of the same streets, and
inference over scenes asks whether it would hold for new streets. We report the
bootstrap over scenes with every comparison between models. The $p$ values of
these comparisons are not corrected for multiple comparisons. Of the nine
comparisons tested over scenes, the two that are significant remain so after
Bonferroni correction.

Each comparison was also run on the pre-stimulus window, which passes through the
same estimation but contains no response to the image. It provides a baseline for
the correspondence values.

To ask whether two feature spaces capture the same part of the neural geometry,
we computed the partial Spearman correlation between one model RDM and each
participant's neural RDM with the other model RDM partialled out. Their joint
correspondence is that of the equally weighted sum of the two rank-transformed
model RDMs, which involves no fitting.

For the depth analysis, the relative depth of an extracted layer is its index
divided by the total depth of the encoder, so that the projected embedding of a
transformer and the logits of a ResNet both have a depth of 1. The peak latency
of a layer is the centre of mass of its correspondence time course after 100\,ms,
restricted to the part above half of its maximum. This is more stable than the
latency of the maximum itself. The analysis includes the eleven models with at
least three extracted layers of defined depth.

\subsection{Dependence on the evaluative goal}

A vision model returns one embedding per image whatever the viewer intends. To
test whether the neural geometry is also independent of the viewer's goal, we
used the fact that every scene was viewed once before each of the nine ratings.
For a given scale we split the participants into two halves, averaged the
whitened patterns of each scene within each half, and computed one RDM per half.
Let $w_i$ be the correlation between the two half-sample RDMs for the same scale
$i$, which estimates the reliability of the geometry under that scale, and
$b_{ij}$ the correlation between the RDM of one half for scale $i$ and the RDM of
the other half for scale $j$. The scales differ in reliability, so $b_{ij}$
cannot be compared with $w_i$ directly. We instead report
\begin{equation}
R \;=\; \underset{i \neq j}{\mathrm{mean}}\;
        \frac{b_{ij}}{\sqrt{w_i\,w_j}},
\label{eq:taskset}
\end{equation}
which divides each cross-scale similarity by the geometric mean of the two
reliabilities involved. If the geometry does not depend on the evaluative goal,
cross-scale similarity is limited only by reliability and $R=1$. A value reliably
below 1 indicates that the goal changes the geometry. The two halves contain
different participants, so noise shared between trials of the same participant
enters neither quantity. We repeated the analysis over 20 random splits of the
participants and averaged the similarities over splits before forming $R$. The
patterns of each block were centred as for the main RDMs. Appendix~\ref{app:taskset}
validates the measure on synthetic data.

\subsection{Appraisal prediction}
\label{sec:readout-methods}

To relate neural correspondence to the way these representations are used in
urban analytics, we predicted the nine group-mean appraisal ratings from each
feature space by ridge regression with 11-fold cross-validation over scenes. The
ridge penalty was selected within each training set by efficient leave-one-out
cross-validation over 25 values between $10^{-2}$ and $10^{6}$. Each scale was
predicted separately and scored by the Pearson correlation between predicted and
observed ratings over all held-out scenes. The reported $r$ is the mean of the
nine correlations. Table~\ref{tab:models} gives $r$ for the layer that best
matches the neural geometry and for the layer of the same feature space that
predicts the ratings best. Figure~\ref{fig:readout} uses the former, and the
comparison across models is reported for both. Appendix~\ref{app:reweight}
describes a further analysis in which the features were reweighted towards the
neural geometry before the ratings were predicted.

\section{Results}
\label{sec:results}

\begin{figure}[t]
\centering
\includegraphics[width=\linewidth]{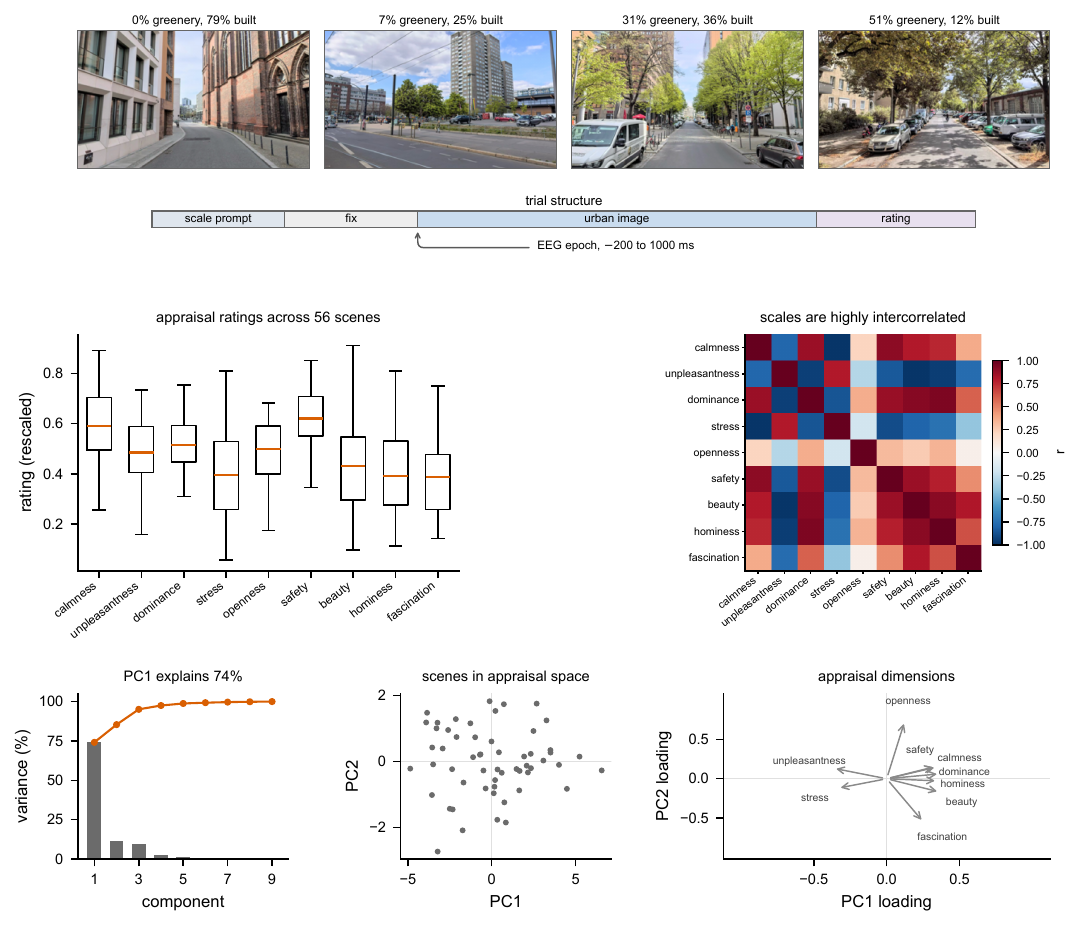}
\caption{Stimuli, task and the structure of urban appraisal. Top: four scenes
chosen to span the measured composition of the set, labelled with the share of
pixels segmented as vegetation and as building, and the trial sequence. A prompt
names the scale to be rated before the image appears, so each of the nine
presentations of a scene is made under a different evaluative goal. Middle:
distribution of the nine ratings across scenes, and their intercorrelation.
Bottom: the nine group-level appraisal measures are largely summarised by two
principal components.}
\label{fig:paradigm}
\end{figure}

\subsection{The data contain reliable scene-specific structure}
\label{sec:res-reliable}

Ratings were consistent across participants. The split-half reliability of the
group-mean rating profile, corrected with the Spearman--Brown formula, ranged from
0.94 to 0.98 across the nine scales. The scales were strongly intercorrelated,
and two principal components accounted for 85\% of the variance across scenes.
The first (74.1\%) contrasted scenes rated calm, safe, beautiful and homely with
scenes rated stressful and unpleasant, and the second (11.2\%) separated openness
from fascination (Figure~\ref{fig:paradigm}). The appraisal ratings are thus
reliable and low-dimensional.

\begin{figure}[t]
\centering
\includegraphics[width=\linewidth]{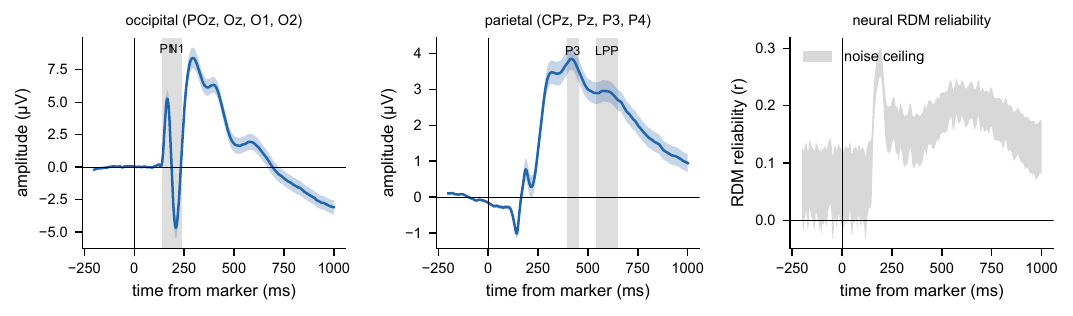}
\caption{Occipital and parietal grand averages with the P1, N1, P3 and LPP
windows shaded, and the reliability of the time-resolved neural RDMs. Times are
relative to the stimulus marker (Section~\ref{sec:preproc}). Shaded bands show
the standard error over participants and the noise-ceiling bounds respectively.}
\label{fig:validation}
\end{figure}

The event-related potentials agree with those reported for this dataset with a
different preprocessing pipeline \citep{zaehme2026perception}. The four
components appeared at the expected sites: an occipital P1 ($+3.67\,\mu$V,
$t_{62}=6.35$, $p=2.8\times10^{-8}$) and N1 ($-3.26\,\mu$V, $t_{62}=-4.08$,
$p=1.3\times10^{-4}$), and a parietal P3 ($+3.78\,\mu$V, $t_{62}=12.14$) and LPP
($+2.90\,\mu$V, $t_{62}=9.69$; both $p<10^{-13}$). This analysis includes all 63
participants. The relation of the components to scene content was also
reproduced. P1 amplitude increased with vegetation ($r=0.43$, $p=0.001$), and the
N1 became more negative as straight-edge density increased ($r=-0.52$,
$p<0.001$), in line with the original report (Figure~\ref{fig:validation}).

Cross-validated dissimilarities between scenes were close to zero before the
image appeared and rose steeply afterwards, from a mean crossnobis distance of
0.007 in the pre-stimulus window to 1.62 between 150 and 300\,ms. Agreement
between participants followed the same course. At single time points, the lower
bound of the noise ceiling was 0.000 between 50 and 150\,ms, 0.159 between 150
and 300\,ms, and between 0.12 and 0.14 thereafter. The window-level RDM
(150--500\,ms) used for the model comparison has a lower bound of 0.193. Reliable
scene structure is thus present from about 150\,ms after the marker.

\begin{figure}[t]
\centering
\includegraphics[width=\linewidth]{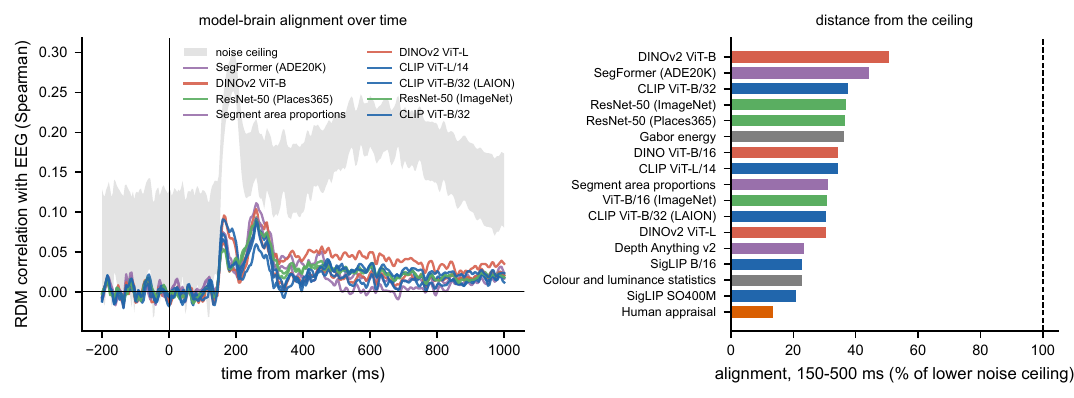}
\caption{Correspondence between model and neural representational geometry.
Left: the best-matching layer of each model at each time point, against the
time-resolved noise ceiling (grey). Right: correspondence for the window-level
RDM over 150--500\,ms as a percentage of the lower bound of that window's
ceiling, coloured by training signal; these are the values in
Table~\ref{tab:models}.}
\label{fig:timecourses}
\end{figure}

\subsection{Vision representations reach at most about half of the noise ceiling}
\label{sec:res-models}

\begin{table}[t]
\centering
\small
\setlength{\tabcolsep}{4pt}
\caption{Correspondence between the representational geometry of each feature space and the neural geometry of urban scenes in the 150--500\,ms window. $\rho$ is the Spearman correlation between the model RDM and each participant's neural RDM, averaged over participants, for the layer that best matches the group-mean neural RDM. The interval is a 95\% bootstrap interval over scenes. \%\,ceiling expresses $\rho$ as a percentage of the lower bound of the noise ceiling (0.193). $\rho_{\text{pre}}$ is the same quantity in the pre-stimulus window, which passes through the same estimation but contains no response to the image. Appraisal $r$ is the cross-validated prediction of the nine appraisal ratings on held-out scenes, from the layer listed (layer) and from the layer of the same feature space that predicts the ratings best (best). The best column is an optimistic estimate, because the layer is chosen on the same held-out predictions.}
\label{tab:models}
\begin{tabular}{llrrrrrr}
\toprule
 & & & & & & \multicolumn{2}{c}{Appraisal $r$} \\
\cmidrule(lr){7-8}
Model & Layer & $\rho$ & 95\% CI (scenes) & \%\,ceiling & $\rho_{\text{pre}}$ & layer & best \\
\midrule
\multicolumn{8}{l}{\textit{language-supervised}} \\
\quad CLIP ViT-B/32 & \texttt{L08} & 0.073 & [0.037, 0.109] & 38 & -0.004 & 0.64 & 0.80 \\
\quad CLIP ViT-L/14 & \texttt{L16} & 0.066 & [0.032, 0.104] & 34 & -0.007 & 0.65 & 0.84 \\
\quad CLIP ViT-B/32 (LAION) & \texttt{L05} & 0.059 & [0.022, 0.098] & 31 & -0.000 & 0.61 & 0.83 \\
\quad SigLIP B/16 & \texttt{L01} & 0.044 & [0.007, 0.085] & 23 & -0.007 & 0.04 & 0.82 \\
\quad SigLIP SO400M & \texttt{embed} & 0.040 & [0.008, 0.074] & 21 & -0.004 & 0.87 & 0.87 \\
\multicolumn{8}{l}{\textit{self-supervised}} \\
\quad DINOv2 ViT-B & \texttt{L08} & 0.098 & [0.064, 0.134] & 51 & -0.002 & 0.76 & 0.81 \\
\quad DINO ViT-B/16 & \texttt{L08} & 0.066 & [0.029, 0.108] & 34 & 0.002 & 0.78 & 0.78 \\
\quad DINOv2 ViT-L & \texttt{L16} & 0.059 & [0.024, 0.097] & 30 & -0.001 & 0.75 & 0.80 \\
\multicolumn{8}{l}{\textit{category-supervised}} \\
\quad ResNet-50 (ImageNet) & \texttt{layer4} & 0.071 & [0.035, 0.109] & 37 & 0.007 & 0.69 & 0.71 \\
\quad ResNet-50 (Places365) & \texttt{layer3} & 0.071 & [0.037, 0.107] & 37 & 0.002 & 0.75 & 0.79 \\
\quad ViT-B/16 (ImageNet) & \texttt{L08} & 0.059 & [0.015, 0.104] & 31 & -0.008 & 0.64 & 0.69 \\
\multicolumn{8}{l}{\textit{dense prediction}} \\
\quad SegFormer (ADE20K) & \texttt{layout} & 0.085 & [0.048, 0.125] & 44 & 0.002 & 0.57 & 0.75 \\
\quad Segment area proportions & \texttt{ade20k} & 0.060 & [0.023, 0.100] & 31 & -0.007 & 0.57 & 0.57 \\
\quad Depth Anything v2 & \texttt{layout} & 0.045 & [0.009, 0.083] & 23 & -0.002 & 0.30 & 0.30 \\
\multicolumn{8}{l}{\textit{interpretable}} \\
\quad Gabor energy & \texttt{energy} & 0.070 & [0.028, 0.116] & 36 & 0.019 & 0.51 & 0.51 \\
\quad Colour and luminance statistics & \texttt{colour} & 0.044 & [0.008, 0.082] & 23 & 0.003 & 0.49 & 0.49 \\
\multicolumn{8}{l}{\textit{human judgement}} \\
\quad Human appraisal & \texttt{ratings} & 0.026 & [-0.000, 0.060] & 14 & 0.002 & -- & -- \\
\bottomrule
\end{tabular}
\end{table}

All seventeen feature spaces showed a positive correspondence with the neural
geometry in the 150--500\,ms window, and all remained below the noise ceiling
(Table~\ref{tab:models}, Figure~\ref{fig:timecourses}). The highest value was
obtained for DINOv2 ViT-B at layer 8 ($\rho=0.098$, 95\% CI over scenes
$[0.064, 0.134]$), which is 51\% of the lower bound of the noise ceiling and
corresponds to about a fifth of the reliable variance in the neural geometry
(Appendix~\ref{app:norm}). Among the image-based feature spaces the
lowest was 21\% of the ceiling, for SigLIP SO400M. In the
150--300\,ms window, by the end of which at most 12\% of trials contain a large
horizontal saccade (Appendix~\ref{app:window}), the best feature space reaches
50\% of that window's noise ceiling (Table~\ref{tab:robustness}).

The correspondence is tied to the stimulus. The cluster-based permutation test
over time gave at least one significant cluster for every feature space, with
onsets between 136 and 244\,ms. For DINOv2 ViT-B a single cluster extended from
152 to 1000\,ms ($p<0.001$). In the pre-stimulus window, correspondence was
between $-0.008$ and $0.019$, and every feature space exceeded its own
pre-stimulus value ($p \le 8.5\times10^{-3}$). Over scenes, the bootstrap
interval excluded zero for every image-based feature space.

The Gabor energy descriptor reached $\rho=0.070$, or 36\% of the ceiling, the
level of most pretrained models. It was lower than DINOv2 ViT-B over participants
(Wilcoxon $p=0.014$) but not over scenes (bootstrap $p=0.26$). Over participants
it exceeded the two SigLIP models ($p=0.012$ and $0.009$) and did not differ from
the other language-supervised models; over scenes none of these differences was
significant (bootstrap $p \ge 0.31$).

The Gabor descriptor and DINOv2 ViT-B capture largely different parts of the
neural geometry. Their RDMs are only weakly correlated ($\rho=0.13$), and
partialling out the Gabor RDM lowered the correspondence of DINOv2 ViT-B from
0.098 to 0.090, while partialling out the DINOv2 ViT-B RDM lowered that of the
Gabor descriptor from 0.070 to 0.058. With the Gabor RDM partialled out, every
other image-based feature space except Depth Anything V2 (63\%) retained at
least 78\% of its correspondence. The two RDMs combined without fitting reached
$\rho=0.109$, or 57\% of the ceiling.

The geometry of the appraisal ratings corresponded only weakly to the neural
geometry ($\rho=0.026$, 14\% of the ceiling, 95\% CI over scenes
$[-0.000, 0.060]$), less closely than any image-based feature space. With ratings
standardised across scenes and Euclidean distance, which does not depend on the
range or direction of a scale, the correspondence was $\rho=0.028$. In this time
window, scenes that are appraised alike are therefore not, to any large extent,
scenes that evoke similar neural patterns.

Differences between individual models were consistent across participants but
mostly not across scenes. DINOv2 ViT-B, for example, exceeded CLIP ViT-B/32 in
the paired test over participants ($p=0.008$) but not when scenes were resampled
($p=0.13$). We therefore do not interpret the order of models within the panel
beyond the specific contrasts reported. The result that holds over both
participants and scenes is the distance of the whole panel from the noise
ceiling. Correcting the correlations for attenuation, without reference to the
ceiling, gives nearly the same order (rank correlation 0.985) and a similar level
(Appendix~\ref{app:norm}).

\subsection{Correspondence follows layer depth but not model size}

Within a model, deeper layers corresponded to later neural responses. Relative
layer depth was positively correlated with peak latency in 9 of the 11 models
with three or more extracted layers (mean rank correlation $+0.66$,
$t_{10}=4.19$, $p=0.002$). The relation reached $\rho=1.00$ in DINOv2 ViT-L and
$\rho=0.98$ in CLIP ViT-B/32 and ResNet-50 trained on ImageNet ($p=0.005$), and
it was negative, though not significantly, in the two SigLIP models. The
temporal ordering of layers reported for object and scene recognition
\citep{cichy2016comparison,cichy2017dynamics} is thus also found for urban
scenes.

Model size and training domain showed no such relation. In the two families with
matched size variants the larger model was not better aligned: DINOv2 ViT-L
reached 30\% of the ceiling against 51\% for ViT-B, a difference that holds over
scenes (bootstrap $p=0.002$), and SigLIP SO400M 21\% against 23\% for SigLIP
B/16. For the two ResNet-50 models, training on scene photographs (Places365,
36.6\%) gave a similar correspondence to training on object photographs
(ImageNet, 37.1\%). These are only three matched comparisons, but in none of them
did a larger model or scene-specific training data increase correspondence.

\begin{figure}[t]
\centering
\includegraphics[width=\linewidth]{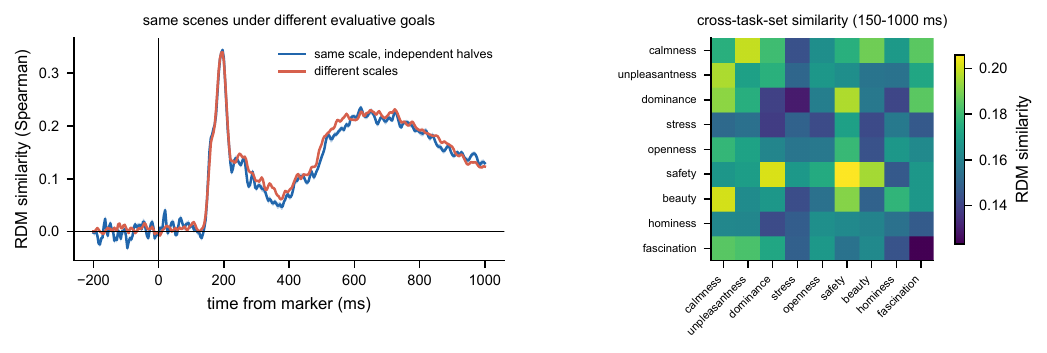}
\caption{No detectable dependence of the neural geometry on the evaluative goal.
Left: RDM similarity between independent halves of the participants rating the
same scale
and rating different scales. Right: cross-task similarity for every pair of
scales. Shaded bands show the standard error over 20 random participant splits.}
\label{fig:taskset}
\end{figure}

\subsection{The neural geometry is stable across evaluative goals}

If the neural geometry of a scene changed with the judgement the viewer was about
to make, a model with one fixed embedding per image could match it only in part.
We found no such change. After correction for reliability, the geometry was as
similar between halves of the participants rating different scales as between
halves rating the same scale: the ratio $R$ of Equation~\ref{eq:taskset} was 1.06
over the 150--1000\,ms window and did not fall below 1 in any time window
(Figure~\ref{fig:taskset}). The neural geometry compared with
the models in Table~\ref{tab:models} thus does not detectably depend on the
rating task. Appendix~\ref{app:taskset} discusses the upward bias of $R$ and
reports a second estimator.

\begin{figure}[t]
\centering
\includegraphics[width=\linewidth]{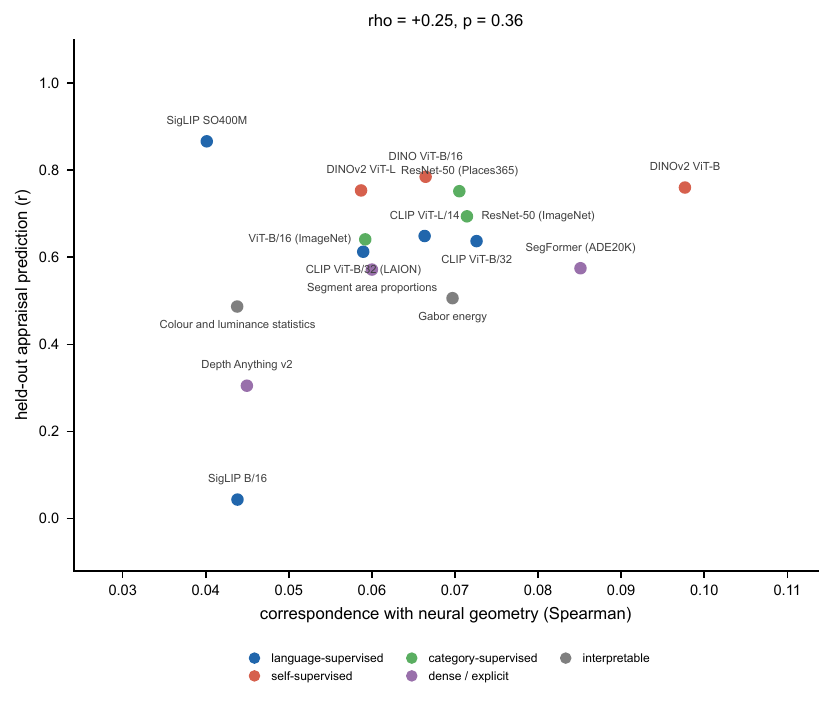}
\caption{Appraisal prediction and neural correspondence across models. Each model
is placed by the correspondence of its most brain-aligned layer, as in
Table~\ref{tab:models}, and by the cross-validated prediction of the nine
appraisal ratings from that layer. The two measures are not reliably related
across models.}
\label{fig:readout}
\end{figure}

\subsection{Appraisal prediction and neural correspondence diverge}

The same feature spaces predict the appraisal ratings well. Ridge regression on
the nine group-mean ratings, evaluated on held-out scenes, reached $r=0.87$ for
SigLIP SO400M and $r=0.84$ for CLIP ViT-L/14 from the layer of each model that
predicts the ratings best (Table~\ref{tab:models}). Seven feature spaces predict
the ratings at $r \ge 0.80$ from their best layer, and their correspondence with
the neural geometry ranges from 21\% to 51\% of the noise ceiling. The best
predictor, SigLIP SO400M, has the lowest correspondence of the image-based
feature spaces, lower than DINOv2 ViT-B also over scenes (bootstrap $p \le 0.001$).
The two best-aligned feature spaces, DINOv2 ViT-B and SegFormer,
predict appraisal at $r=0.76$ and $r=0.57$ from their best-aligned layers, and at
$r=0.81$ and $r=0.75$ from their best-predicting layers. Across the sixteen
image-based feature spaces, appraisal prediction and neural correspondence were
not reliably related, whether prediction was taken from the layer that best
matches the neural geometry ($\rho=0.25$, 95\% CI $[-0.28, 0.66]$;
Figure~\ref{fig:readout}) or from the layer that predicts the ratings best
($\rho=-0.05$, 95\% CI $[-0.53, 0.46]$).

Reweighting the features towards the neural geometry, with weights fitted on
training scenes, lowered appraisal prediction for every model and, in 15 of 16
models, also lowered correspondence with the neural geometry on held-out scenes
(Appendix~\ref{app:reweight}). We therefore treat it as an unsuccessful
manipulation and draw no conclusion from it about the relation between appraisal
and neural correspondence.

\section{Discussion}
\label{sec:discussion}

This study asked whether the vision representations used to model urban appraisal
also arrange street scenes as the brain does. In EEG recorded while people viewed
and rated Berlin streets, the best-aligned feature space reached 51\% of the noise
ceiling, and a simple descriptor of oriented edges reached the level of most
pretrained models while capturing a different part of the neural geometry. How well a feature space predicted the ratings was not reliably
related to how closely it corresponded to the neural geometry.

\subsection{Prediction and representation come apart}

Accurate prediction of appraisal did not come with close correspondence to the
neural scene geometry. The seven feature spaces that predicted the ratings at
$r \ge 0.80$ reached between 21\% and 51\% of the noise ceiling. The best
predictor had the lowest correspondence of the image-based feature spaces, and
across models the two measures were not reliably related. A likely reason is that the
two measures depend on different content. Appraisal ratings depend on what a
scene contains and what it means to the viewer, and language supervision yields
embeddings that are organised by such semantic content
\citep{radford2021learning}. The neural geometry in the 150--500\,ms window does
not follow this organisation, as two observations from the same participants
show.

The geometry of their own ratings corresponded to the neural geometry less closely
than any image-based feature space (Table~\ref{tab:models}), whether the ratings
were used as coded or standardised. In the event-related potentials, the early
occipital components followed image content and were not associated with
the ratings, which were associated with the later parietal P3 and LPP on some of
the scales \citep{zaehme2026perception}. Appraisal and the early neural response
to a scene are thus organised differently in people as well, and a model fitted
to one need not resemble the other.

Predicting an appraisal outcome and reproducing the representation of the scene
are therefore two separate targets of evaluation. A model can meet the first
without meeting the second, and improvement on the first gives no assurance about
the second.

\subsection{Implications for urban analytics and planning}

For studies whose aim is to estimate an appraisal outcome, such as mapping
perceived safety across a city or relating perceived beauty to walking or housing
prices, held-out predictive accuracy remains the appropriate criterion, and the
representations tested here meet it. Our results give no reason to doubt such
maps on the grounds that the underlying model represents the street differently
from people.

The results matter for uses that go beyond the predicted score. The first is
similarity. Grouping streets into visual types and measuring how alike two
neighbourhoods or cities look both operate on distances between embeddings
\citep{wang2020urban2vec,huang2025measuring,guo2025urban,verma2025assessing}.
These distances are informative about perceived similarity only if the embedding
arranges scenes as viewers do, which predictive accuracy does not establish. The
second is explanation. Feature-importance and
attribution scores are often read as showing which street elements make a place
feel safe or beautiful, and are then translated into design recommendations
\citep{liu2023interpretable,guo2025deep}. What they show is
which elements the model relies on. Since the model reaches the rating through a
representation that differs from the neural one, these need not be the elements
that drive the human response. \citet{kang2026decoding} tested such explanations
against human data. They compared seven attribution methods for a model of
perceived safety with the eye movements of 127 people choosing which of two
street scenes looked safer, and the methods differed in how closely they matched
human gaze.

The third use is the assessment of design proposals, in which a perception model
scores rendered or generated streetscapes before anything is built
\citep{joglekar2020facelift,hu2025ursimulator}. Such images alter specific
elements of a scene, and a model that reproduces average ratings of existing
streets through a differently organised representation may respond to those
alterations differently from people. Our data do not test this case directly.

Two practical steps follow. When a study interprets embedding distances or model
explanations as perceptual, the interpretation should be supported by human data
for a sample of the scenes, such as similarity judgements
\citep{garridovalenzuela2026pixels}, eye movements \citep{kang2026decoding} or
neural recordings, in the same way that predicted scores are validated against
held-out ratings. In addition, new representations for urban analytics, including
models trained specifically on street-level imagery, can be tested for neural
correspondence at low cost with the comparison introduced here, since it
requires only their embeddings of 55 images.

\subsection{Model size, supervision and layer depth}

The panel covers four kinds of training objective, image encoders from 26 to 428
million parameters, and training sets from 1.3 million images to billions of
image--text pairs. In the three matched comparisons available, neither a larger
model nor training on scene photographs increased correspondence, and in the
DINOv2 pair the larger model was less aligned. This agrees with comparisons
against human similarity judgements \citep{muttenthaler2023human} and against
fMRI \citep{conwell2024large}, in which model size did not improve alignment.
Both studies found larger effects of training data, which our single matched
comparison of training data does not show.

A descriptor of oriented edges reached the level of most pretrained models and
captured a part of the neural geometry largely different from that captured by
the best of them:
partialling out the descriptor lowered the correspondence of DINOv2 ViT-B by 8\%,
partialling out DINOv2 ViT-B lowered that of the descriptor by 17\%, and most
other feature spaces retained at least 78\% of their correspondence when the
Gabor geometry was partialled out (Section~\ref{sec:res-models}). What current
models share with the scalp-recorded geometry is therefore largely not the
oriented edge energy that the descriptor captures, and the best of them misses
most of that part. Combined without fitting, the best model and the descriptor
reached 57\% of the noise ceiling, so a substantial part of the reliable neural
geometry is captured by neither.

Layer depth behaved differently from model size. Deeper layers corresponded to
later neural responses in 9 of 11 models, as found for object and scene
recognition \citep{cichy2016comparison,cichy2017dynamics}. The models thus
reproduce the temporal order of visual processing.

\subsection{No detectable dependence of the neural geometry on the evaluative goal}

Every scene was viewed before nine different ratings, which allowed us to test
whether the neural geometry depends on the judgement the viewer is about to make.
\citet{zaehme2026perception} averaged over the nine presentations and left the
effect of the cue open. At the level of scene geometry within the first second we
found no detectable effect. The incomplete correspondence of fixed image embeddings is
therefore not explained by a neural representation that shifts with the task. A
goal-dependent component no larger than the upward bias of our estimator
(Appendix~\ref{app:taskset}), or one that arises in sources poorly resolved by
EEG, would not be detected by this analysis.

\subsection{Scope of the EEG measurement}

The noise ceiling is estimated from the agreement between participants. It
therefore describes the scene structure that is shared across people and that a
model computed from the image could in principle reproduce, and the distance
between the models and the ceiling is a distance from this shared structure. What
the ceiling contains is limited to what EEG records. Scalp potentials mainly
reflect synchronous activity in superficial cortex, so representations carried by
sparse or deep neural populations contribute little to the measured geometry and
to the ceiling. Our conclusions concern the part of the scene representation that
EEG resolves.

\subsection{Limitations}

The stimulus set consists of 55 scenes from a single city. It was assembled to
contrast streets that differ in greenery and building height, and it does not
sample streets representatively: vegetation is effectively absent from 26 of the
55 scenes and covers more than a third of the frame in only a few. The results
should not be generalised to other cities or to street types absent from the set.
The small number of scenes also limits what can be inferred. Under the bootstrap
over scenes the distance of the whole panel from the ceiling is stable, whereas
the order of models within the panel is not, and a decomposition of the neural
geometry into visual, semantic, spatial and appraisal components could not be
estimated (Appendix~\ref{app:decomp}).

The scenes were divided between the two blocks in the same way for every
participant, so block and scene are confounded by design, and the correction for
this confound assumes that the difference between the blocks is additive.

Each image was shown for 3000\,ms, which is long enough for several fixations,
and no electro-oculogram was recorded. Estimated from the bipolar F8--F7 signal,
horizontal eye movements had begun in a quarter to a half of the trials by the end
of the main window and in most trials by 1000\,ms (Appendix~\ref{app:window}).
The later part of the main window and the rest of the epoch therefore include
the effects of eye movements, and we cannot remove this contribution.

Ratings were collected once after each image, so the appraisal geometry describes
the outcome of the judgement, and the process leading to it is not captured.
Finally, participants viewed static photographs on a screen. Whether the same
relations hold while people move through a real street is a question for mobile
EEG.

\section{Conclusion}
\label{sec:conclusion}

We compared seventeen feature spaces with the neural geometry of Berlin street
scenes, estimated from the EEG of adults who viewed and rated them, and scaled
every comparison by the structure that is shared across participants. No feature
space reached more than about half of the noise ceiling, and a Gabor energy
descriptor reached the level of most pretrained models while capturing a largely
different part of the neural geometry. In the two model families with size variants the larger
model was not better aligned, while deeper layers corresponded to later neural
responses. The same representations predicted held-out appraisal ratings well,
and how well a model predicted appraisal was not reliably related to how closely
it matched the neural geometry.

For urban analytics these results separate two uses of an embedding. Held-out
predictive accuracy remains the right criterion for a study that estimates
perceived safety or beauty, and the representations tested here meet it. It does
not justify reading distances or dimensions of the embedding as properties of
human perception, which requires direct evidence from human similarity data. The
comparison reported here can supply part of this evidence for any new
representation, including models trained on street-level imagery, from its
embeddings of 55 images.

\section*{Data and code availability}

All inputs to this study are public. The EEG recordings are dataset
\texttt{ds006850} on OpenNeuro, released under CC0
(\url{https://doi.org/10.18112/openneuro.ds006850.v1.0.0}). The stimuli and
experiment code are in the \texttt{urban\_\allowbreak appraisal-experiment}
repository of the original authors, and the ratings, segmentation maps and
low-level descriptors are in their
\texttt{perception2appraisal\_\allowbreak analyses} repository. Model weights were
taken from their official releases. Our code, which downloads these inputs and
reproduces every figure and table, will be made publicly available on
publication.

\bibliographystyle{apalike}
\bibliography{main}

\newpage
\appendix
\section{Preprocessing details}
\label{app:preproc}

\paragraph{Channels.} The 66 recorded channels comprise 64 scalp electrodes, ECG
and electrodermal activity. The last two were removed before analysis.

\paragraph{Bad-channel criterion.} An electrode was marked as bad when it was
flat, when its largest absolute correlation with any other electrode was below
$0.4$, or when its standard deviation exceeded eight times the montage median. We
did not use variance alone, because blinks give working frontopolar electrodes a
high variance, and a variance criterion marked these electrodes in almost every
participant. Scalp potentials are spatially smooth, so a working electrode
correlates with its neighbours whatever artefacts it carries, whereas a
disconnected electrode does not.

\paragraph{Reading the public release.} Two properties of the released files
affect any reanalysis. The BIDS headers do not declare channel types and the
stored values are in microvolts, so they must be multiplied by $10^{-6}$ before
any amplitude-based step. The marker for the beauty scale contains a non-ASCII
character, so marker strings must be normalised to ASCII before matching,
otherwise the beauty trials are lost.

\paragraph{Trial timing.} The original report gives 500\,ms for the fixation
cross. In the released marker streams the stimulus marker follows the onset of
the fixation cross by a median of 1.02\,s and the onset of the word pair by a
median of 2.04\,s (31{,}749 trials), so we describe the fixation period as about
1000\,ms.

\section{Verification of the dissimilarity estimator}
\label{app:crossnobis}

The crossnobis estimator in Equation~\ref{eq:crossnobis} has an expected value of
zero when conditions do not differ. In 200 simulations with 10 conditions, 20
channels, 8 trials per condition and no condition structure, the mean estimated
dissimilarity was $0.028$ (SEM $0.041$, $t=0.68$), which is consistent with
zero. When condition means were added to the simulated data, the same code
returned large positive distances.

\section{Noise ceiling bias}
\label{app:ceiling}

The upper bound of the noise ceiling correlates each participant's RDM with a
group mean that includes that participant. It is therefore positively biased when
RDMs are unreliable and approaches $1/\sqrt{n}$ for pure noise. With $n=61$ this
gives $0.128$. In the pre-stimulus window the measured upper bound averages
$0.128$ and the lower bound $0.000$, as expected when the RDMs contain no
stimulus-specific structure. For this reason we express model performance
relative to the lower bound throughout.

\section{Correction for attenuation as an alternative to the noise ceiling}
\label{app:norm}

The percentage of the noise ceiling is a ratio of two correlations, and its value
depends on the bound used as the denominator. As a check that does not involve
the ceiling, we corrected the correlation with the group-mean RDM for
attenuation. The group-mean RDM has a split-half reliability of $0.578$, or
$0.732$ after Spearman--Brown correction to the full sample. A model RDM is a
deterministic function of the images and contains no measurement noise, so the
correlation between the model geometry and the noise-free neural geometry can be
estimated as $\rho_{\text{true}} = \rho_{\text{obs}} / \sqrt{0.732}$. This
estimate is bounded by one and does not depend on the number of participants.

The two normalisations agree closely, with a rank correlation of $0.985$ across
the panel ($p=6.3\times10^{-13}$). DINOv2 ViT-B correlates with the group-mean
geometry at $\rho=0.390$, or $0.456$ after correction, compared with 51\% of the
ceiling in the main text. The percentage of the ceiling is the higher of the two
figures, as expected when dividing by a lower bound. Squaring the corrected
correlations, the best model accounts for 21\% of the reliable variance in the
neural geometry, the image-based feature spaces for between 3.4\% and 21\%, and
the appraisal ratings for 1.3\%.

\section{Layer selection}
\label{app:selection}

The layer reported for each model is the one that best matches the group-mean
RDM, to which every participant contributes. When the layer was instead selected
from 60 participants and evaluated on the remaining one, the selected layer was
the same in every fold for all 12 multi-layer models.

\section{Validation of the goal-dependence measure}
\label{app:taskset}

We validated the ratio $R$ of Equation~\ref{eq:taskset} on synthetic data with a
known amount of goal-dependent modulation and the same numbers of participants,
scenes, scales and channels as the real data. With no modulation, $1-R$ was
$-0.015$. With a moderate goal-specific component it was $0.558$, and with a
strong one $0.913$. The measure thus detects modulation when it is present and
returns a value close to zero when it is absent. Because $R$ divides by noisy
reliability estimates, it is biased slightly upwards, as the simulation without
modulation shows, and the bias is expected to grow as reliability falls. In the
real data $R$ was highest in the 300--500\,ms window (1.22), where the reliability
within scales was lowest ($0.084$). The ratio of the mean cross-scale similarity to
the mean within-scale similarity divides only once and is less affected by noisy
denominators. It was 1.05 over 150--1000\,ms and 1.17 in the 300--500\,ms window.
Goal dependence would lower either estimate, and neither fell below 1 in any
window. An upward bias of the size observed could mask a goal-dependent component
of similar magnitude.

\begin{figure}[t]
\centering
\includegraphics[width=\linewidth]{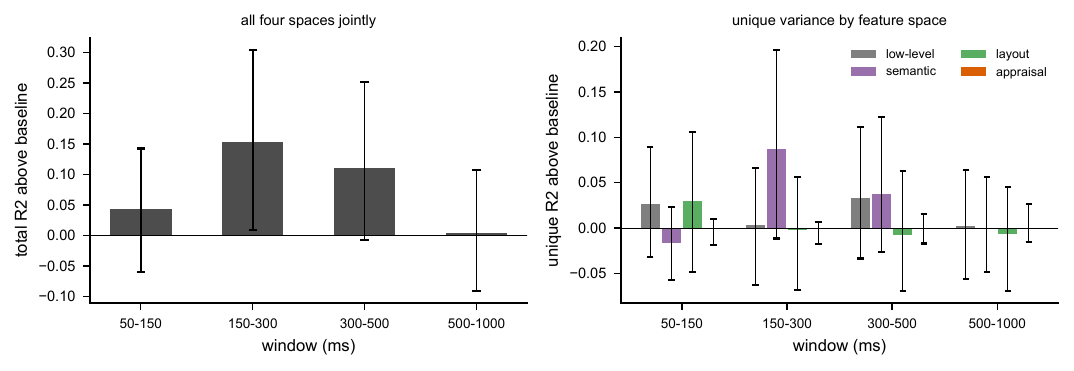}
\caption{Decomposition of the group neural RDM into early visual, semantic,
spatial-layout and appraisal components, expressed relative to the same analysis
run on the pre-stimulus window. Error bars are 95\% intervals from resampling
scenes. The intervals of the unique contributions include zero in every window.}
\label{fig:variance}
\end{figure}

\section{Decomposition of the neural geometry}
\label{app:decomp}

We attempted to decompose the group-mean neural RDM into early visual, semantic,
spatial-layout and appraisal components by regression on eight model RDMs in
four groups: Gabor energy, colour and luminance statistics and the low-level
descriptors released with the dataset (early visual); segment area proportions
and the SegFormer layout (semantic); the Depth Anything layout and the depth
statistics released with the dataset (spatial layout); and the appraisal ratings.
The unique variance of a group is the loss in $R^2$ when that group is removed. A
regression with several correlated predictors gives a positive $R^2$ on a single
noisy RDM even without signal ($R^2=0.049$ in the pre-stimulus window), so every
quantity is expressed relative to the same analysis on the pre-stimulus window.

The total explained variance above baseline was $0.15$ between 150 and 300\,ms
(95\% interval from resampling scenes $[0.01, 0.30]$), and its interval included
zero in the other windows. The unique contribution of every group had an
interval that included zero in every window (Figure~\ref{fig:variance}). With 55
scenes and eight correlated predictors the decomposition is not identifiable,
and we do not interpret it.

Inference over participants and inference over scenes answer different
questions. The first asks whether an effect would recur with new viewers of these
55 streets, and the second whether it would recur with new streets, which is what
a general claim about urban perception requires. The two can give different
answers, as the comparison between models in Section~\ref{sec:res-models} shows.

\begin{figure}[t]
\centering
\includegraphics[width=\linewidth]{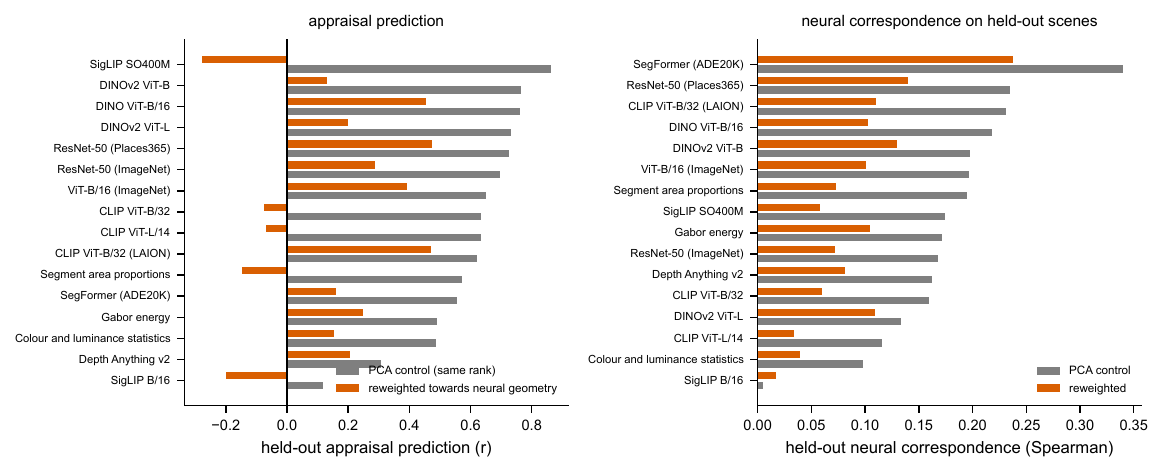}
\caption{Reweighting model features towards the neural geometry. Left: held-out
appraisal prediction falls for every model relative to a control of the same rank
that does not use the neural data. Right: correspondence with the neural geometry
on scenes that contributed to neither the weights nor the PCA basis falls in 15 of
the 16 feature spaces. The reweighting lowers both measures, which indicates
overfitting.}
\label{fig:reweight}
\end{figure}

\section{Reweighting the features towards the neural geometry}
\label{app:reweight}

We tested whether moving a representation towards the neural geometry changes its
prediction of appraisal. Following feature-reweighted RSA
\citep{kaniuth2022feature}, we fitted non-negative weights on the features by
least squares so that the reweighted model RDM approximated the neural RDM, using
the training scenes only. An embedding with several thousand dimensions cannot be
reweighted dimension by dimension from about a thousand training pairs, so each
feature space was first reduced to 40 principal components estimated on the
training scenes. The reweighted features are compared with a control that applies
the same reduction without the neural data, which separates the effect of the
weighting from the effect of the reduction. Each of the 16 image-based feature
spaces enters this analysis once, with its most brain-aligned layer, and the
effect of reweighting is tested across them.

Relative to the control, reweighting lowered the mean prediction of the ratings
from $r=0.60$ to $r=0.15$, and it did so for every one of the 16 feature spaces
($t_{15}=-6.62$, $p=8.2\times10^{-6}$; Wilcoxon $p=3.1\times10^{-5}$). The
control predicted the ratings as well as the full feature set ($r=0.601$ against
$0.602$, $t_{15}=-0.28$, $p=0.78$), so the loss is due to the weighting itself.

Taken alone, this result is ambiguous. It would be informative if the same
weights increased correspondence with the neural geometry, and uninformative if
they only removed variance from the features. We therefore fitted the weights on
half of the scenes and measured correspondence on the other half, which
contributed to neither the weights nor the PCA basis. The reweighted features
were less aligned than the control ($\rho=0.092$ against $0.175$) and improved on
it in only 1 of the 16 feature spaces ($t_{15}=-9.14$, $p=1.6\times10^{-7}$;
Figure~\ref{fig:reweight}). With 55 scenes and 1485 scene pairs the fit does not
recover a neural subspace that generalises, and we treat the manipulation as
unsuccessful.

\section{Robustness to analysis choices}
\label{app:robust}

\begin{table}[t]
\centering
\small
\caption{Sensitivity of the model comparison to analysis choices. Each row repeats the comparison with a different dissimilarity estimator, latency window, participant set or electrode subset; the window is 150--500\,ms unless stated. All rows except the main analysis include the two participants with incomplete scene coverage, and rank $\rho$ compares the order of models in each row with that of the first row, which differs from the main analysis only in this respect. The mean pattern of each block is removed in every row (Section~\ref{sec:geometry}), and the within-block row correlates only pairs of scenes shown in the same block, separately for each block. Best $\rho$ is the highest correspondence in the row, \%\,ceiling expresses it as a percentage of the lower bound of the noise ceiling estimated for the same variant, and the last column names the feature space that reaches it.}
\label{tab:robustness}
\begin{tabular}{lrrrrl}
\toprule
Variant & $n$ & Rank $\rho$ & Best $\rho$ & \%\,ceiling & Best-ranked space \\
\midrule
Crossnobis (reference) & 63 & +1.000 & 0.097 & 51 & DINOv2 ViT-B \\
Crossnobis, 150--300 ms & 63 & +0.882 & 0.116 & 50 & DINOv2 ViT-B \\
Crossnobis, 500--1000 ms & 63 & +0.588 & 0.046 & 27 & DINOv2 ViT-B \\
Crossnobis, within-block pairs & 63 & +0.975 & 0.091 & 49 & DINOv2 ViT-B \\
Correlation distance & 63 & +0.434 & 0.063 & 29 & ResNet-50 (ImageNet) \\
Crossnobis, main analysis & 61 & +0.988 & 0.098 & 51 & DINOv2 ViT-B \\
Correlation distance, posterior electrodes & 63 & +0.645 & 0.104 & 37 & ResNet-50 (ImageNet) \\
Correlation distance, central electrodes & 63 & +0.370 & 0.034 & 24 & SegFormer (ADE20K) \\
Correlation distance, frontal electrodes & 63 & +0.525 & 0.031 & 15 & ResNet-50 (Places365) \\
\bottomrule
\end{tabular}
\end{table}

Without the correction for the block structure (Section~\ref{sec:geometry}), an
RDM that only separates the two blocks corresponded to the neural geometry of the
main-analysis participants at $\rho=0.22$ between 150 and 500\,ms (78\% of the
noise ceiling) and at $\rho=0.16$ between 500 and 1000\,ms (68\%); with the
correction, at $\rho=-0.01$ and $0.00$. The correction lowered the lower bound of
the noise ceiling in the main window from 0.281 to 0.193.

Table~\ref{tab:robustness} repeats the model comparison with a different
dissimilarity estimator, latency window, participant set, set of scene pairs or
electrode subset. Except for the main analysis, every row includes the two
participants with incomplete scene coverage, and its order of models is compared
with the crossnobis analysis of the same 63 participants. Excluding the two
participants, which gives the main analysis, leaves the order almost unchanged
(rank $\rho=0.99$). Restricting the comparison to pairs of scenes from the same
block leaves it almost unchanged ($\rho=0.98$), with the same best feature space
at 49\% of the ceiling.
Restricting the window to 150--300\,ms, before most exploratory eye movements
(Appendix~\ref{app:window}), gives a similar order ($\rho=0.88$). Moving the
window to 500--1000\,ms lowers the agreement ($\rho=0.59$), although DINOv2 ViT-B
still ranks first. Replacing the crossnobis distance with a correlation distance
on evoked patterns over the whole montage or over posterior, central or frontal
electrodes gives weaker agreement ($\rho$ between $0.37$ and $0.65$) and a
different top-ranked feature space. The order of models is thus not stable
across analysis choices, in line with the bootstrap over scenes, and we do not
interpret it. The distance from the noise ceiling is stable: in every variant the
best feature space reaches between 15\% and 51\% of the lower bound of the noise
ceiling estimated for that variant.

\section{Characterisation of the analysis window}
\label{app:window}

We characterised the end of the main window in two ways that do not involve any
feature space. All times are relative to the stimulus marker.

The first concerns the neural geometry itself. Participants were split into
halves, and the group RDM of one half in each 20-ms bin was correlated with the
group RDM of the other half averaged over 600--1000\,ms, with correction for the
reliability of both. This similarity was between 0.02 and 0.25 from 170 to
330\,ms, first reached 0.9 at 530\,ms (95\% interval over participants
450--590\,ms) and remained above 0.9 until the end of the epoch. The geometry thus
changes over the first few hundred milliseconds of the response, so that its
early and late configurations are nearly unrelated, and then settles into a
configuration that persists.

The second concerns eye movements. No electro-oculogram was recorded, so the
bipolar F8--F7 signal of the raw recordings served as a proxy for horizontal eye
position. A trial counted as having left fixation once this signal deviated from
its pre-stimulus mean by more than $k$ times its pre-stimulus standard deviation
for at least 20\,ms. For $k$ between 3 and 5, the signal had left fixation in
4--12\% of trials by 300\,ms, in 26--49\% by 500\,ms and in 64--85\% by
1000\,ms. The proxy is insensitive to small and vertical saccades and can be
affected by lateralised brain activity, so these proportions are approximate.

The 150--500\,ms window therefore covers the period in which the neural geometry
changes and in which most trials do not yet contain a large horizontal saccade.
The later, stable configuration coincides with frequent eye movements, and its
correspondence with the models is reported separately (Table~\ref{tab:robustness}).

\end{document}